[1]**Preetish Kakkar**

[1]**Srijani Mukherjee**

[2] **Hariharan Ragothaman**

[2] **Vishal Mehta**

# Physics Based Differentiable Rendering for Inverse Problems and Beyond

***Abstract: -*** Physics-based differentiable rendering (PBDR) has become an efficient method in computer vision, graphics, and machine learning for addressing an array of inverse problems. PBDR allows patterns to be generated from perceptions which can be applied to enhance object attributes like geometry, substances, and lighting by adding physical models of light propagation and materials interaction. Due to these capabilities, distinguished rendering has been employed in a wider range of sectors such as autonomous navigation, scene reconstruction, and material design. We provide an extensive overview of PBDR techniques in this study, emphasizing their creation, effectiveness, and limitations while managing inverse situations. We demonstrate modern techniques and examine their value in everyday situations.

## I. INTRODUCTION

Integrating physical principles governing light transport, reflection, and refraction sets Physics-based Differentiable Rendering (PBDR) apart from conventional rendering approaches and allows for more realistic simulations of real-world occurrences. This approach makes it possible to simulate light in a novel way, utilizing knowledge from computer graphics, machine learning, and physics. As a result, it is a perfect tool for resolving a wide range of inverse problems.

Recently PBDR has been used in several fields, including scene interpretation, material parameter prediction, and 3D reconstruction. Relying on estimates or pre-defined models, traditional rendering pipelines frequently experienced challenges concerning accuracy and the capacity to deconstruct complicated surroundings. On the other hand, by using gradient facts PBDR makes it easier to optimize scene settings so that they closely match observed data. In computer vision, where inverse rendering has made it capable of reconstructing 3D scene features from 2D images, this skill has shown to be revolutionary [1].

The main idea behind PBDR is to differentiate through the rendering process by utilizing automatic technologies. This enables the approach to produce gradients of an objective function with respect to scene aspects like geometry, material qualities, and lighting which can be further applied to optimization techniques based on gradient descent. The underlying structures can be adjusted thanks to such gradients, which results in inverse problem solutions that are more accurate and effective.

Utilizing methods like differentiable Monte Carlo Integration to effectively generate gradients, recognizable rendering expands over established ideas of light transport theory, such as radiative transfer and Monte Carlo path tracing [3]. Additionally, researchers now have powerful assets at their disposal to examine and apply PBDR approaches because of developments in differentiable rendering frameworks like NVDiffRender and Mitsuba 2 [2].

Our goal in this study is to present a thorough analysis of PBDR techniques, their use in solving inverse issues, and their wider implications for the domains of graphics and machine learning. In addition, we give specific instances that demonstrate the practical applications of PBDR and talk about the experimental setups employed in current research. Finally, we suggest future initiatives to further enhance the scalability and resilience of PBDR methodologies.

## II. RELATED WORK

The origins of differentiable rendering can be found in conventional rendering methods that employed light transport-based algorithms to create images from scene descriptions. The forward challenge of figuring out how light interacts with objects to form an image is solved by classical rendering techniques like path tracing and ray tracing. To retrieve the underlying scene parameters like geometry, material qualities, lighting, etc. from observed images is the inverse problem that interests us in many real-world applications.

[1] IEEE Senior Member

[2] Bayer

[3] athenahealth

[4] Apple

**Differentiable Rendering Frameworks**

Analysis in this field has intensified due to the emergence of frameworks for differentiable rendering. Several important systems that offer tools for effective gradient-based scene parameter optimization have been introduced. One of the most well-known frameworks with the ability to render in both forward and reverse directions is Mitsuba 2. It can compute gradients of intricate light traces with associated scene parameters and helps a variety of rendering approaches, such as differentiable Monte Carlo integration [2].

NVDiffRender is a popular platform that concentrates on real-time GPU-accelerated differentiable rendering [4]. These structures allow investigators to rapidly demonstrate methods for inverse rendering and use them for material recovery, illumination estimation, and 3D shape reconstruction, among other applications.

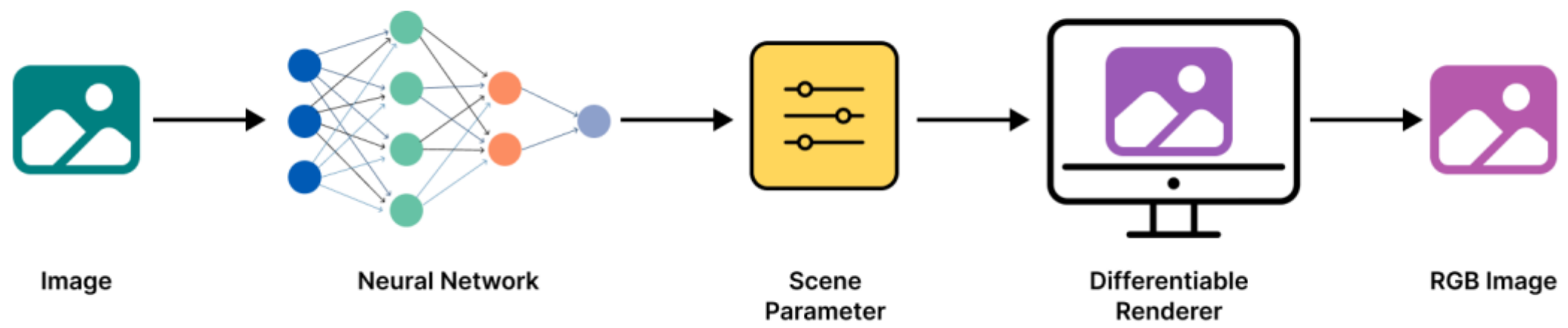

**Applications in Inverse Rendering**

Physically Based Rendering (PBR) structures, which incorporate physically accurate descriptions of light-material associations, represent further breakthroughs in this area. These models can be made recognizable so that tracked image data can be used to optimize material qualities like the parameters of the Bidirectional Reflectance Distribution Function (BRDF) [1].

Differentiable Monte Carlo ray tracing model enables the enhancement of extremely complicated scenes with complex light mobility paths. This method makes use of edge sampling, which maximizes light path parameters for effective gradient computing [4].

Equation 1: Monte Carlo Path Integral for Light Transport

$$L_0(p,\omega_0) = L_e(p,\omega_0) + \int_{\Omega} f_r(p,\omega_i,\omega_0) L_i(p,\omega_i)(\omega_i \cdot n)\, d\omega_i \qquad (1)$$

Where:

$L_0(p,\omega_0)$ is outgoing radiance at point $p$ in direction $\omega_0$, $L_e(p,\omega_0)$ is emitted radiance from the point, $f_r(p,\omega_i,\omega_0)$ is Bidirectional Reflectance Distribution Function (BRDF) which models the material properties, $L_i(p,\omega_i)$ is an incoming radiance at point $p$ from direction $\omega_i$, and $n$ is surface normal at point $p$.

**Differentiable Monte Carlo Integration**

Since Monte Carlo techniques can handle complex lighting interactions like caustics and global illumination, they have been utilized for rendering for a long time. Although there were a few difficulties in managing discontinuous integrals and high variance gradients when trying to make Monte Carlo methods differentiable [5].

Differentiating the light transport equation with regards to scene attributes is necessary for differentiable Monte Carlo integration. The primary difficulty is that light routes can be abrupt or inconsistent, particularly when they involve reflections and refractions. Because of this, scientists have created methods to smooth out these gradients. For example, they can re-parameterize the integrals or concentrate on edge sampling strategies that maximize light parameters for precise gradient flow [4].

**Neural Techniques to Differentiable Rendering**

Neural networks were added to differentiable rendering pipelines with the introduction of deep learning, which has improved their capacity to address inverse problems. A good example is Neural Radiance Fields (NeRF), in which a scattered set of 2D images is used for shaping a neural network to simulate the radiance field of a 3D scene [8]. NeRF can learn a scene's ongoing 3D representation from any perspective. NeRF offers unique view synthesis and precise 3ED scene reconstruction by making the entire pipeline differentiable.

In BRDF estimation, a neural model is trained to forecast material qualities from images, which makes the rendering process very adaptive to different scenarios. Neural networks have also been employed to this task [7].

Equation 2: Neural Radiance Field (NeRF) model

$$C(r) = \int_{t_n}^{t_f} T(t)\sigma\big(x(t)\big)c(x(t), d)dt \qquad (2)$$

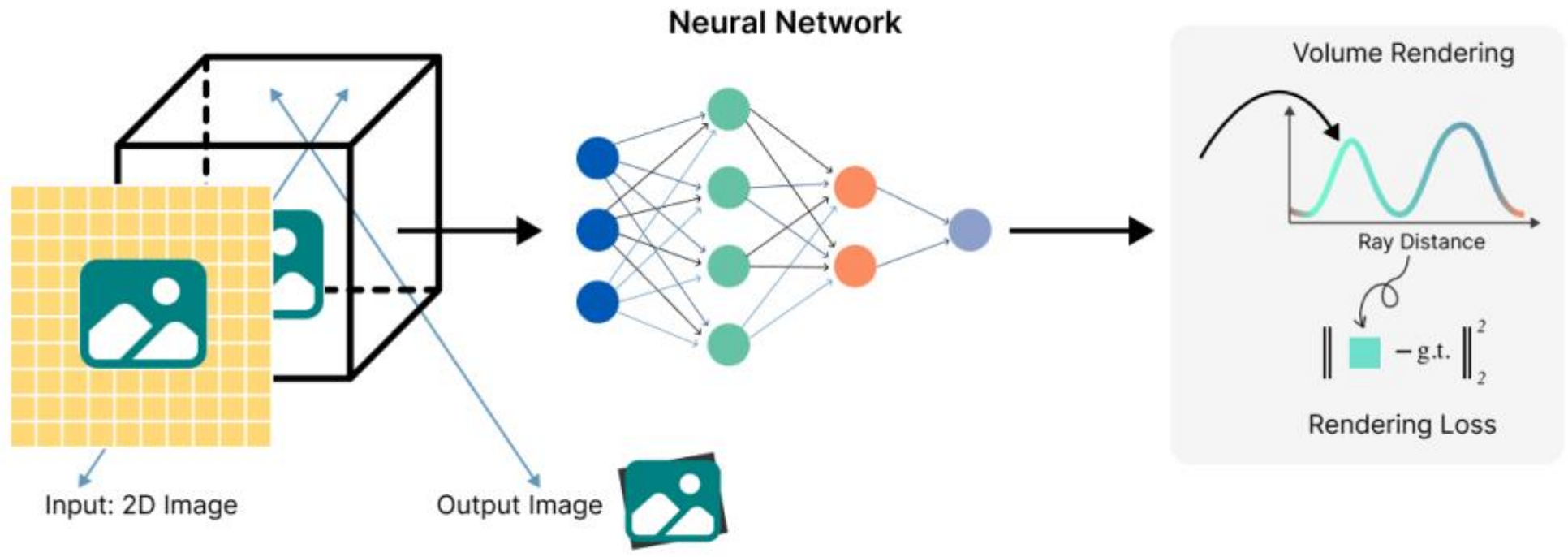

**Figure 2. Neural Radiance Fields Pipeline**

### Challenges and Future Directions

Distinctive rendering shows a lot of potential, but it also has various drawbacks. The computational difficulties of creating high-fidelity scenes with physically realistic light transport models is one major problem. Large quantities of memory and computing power are frequently needed for differentiable rendering pipelines, particularly when working with vast-scale scenes or high-resolution images. The goal of this field's future research should be to make these techniques more scalable.

Managing noisy gradients in Monte Carlo based techniques is another challenge. The optimization process can be significantly impacted by gradient variance, which makes it challenging to converge to precise solutions. Research on methods like variance reduction and adaptive sampling is essential to enhancing the resilience of differentiable rendering techniques [5].

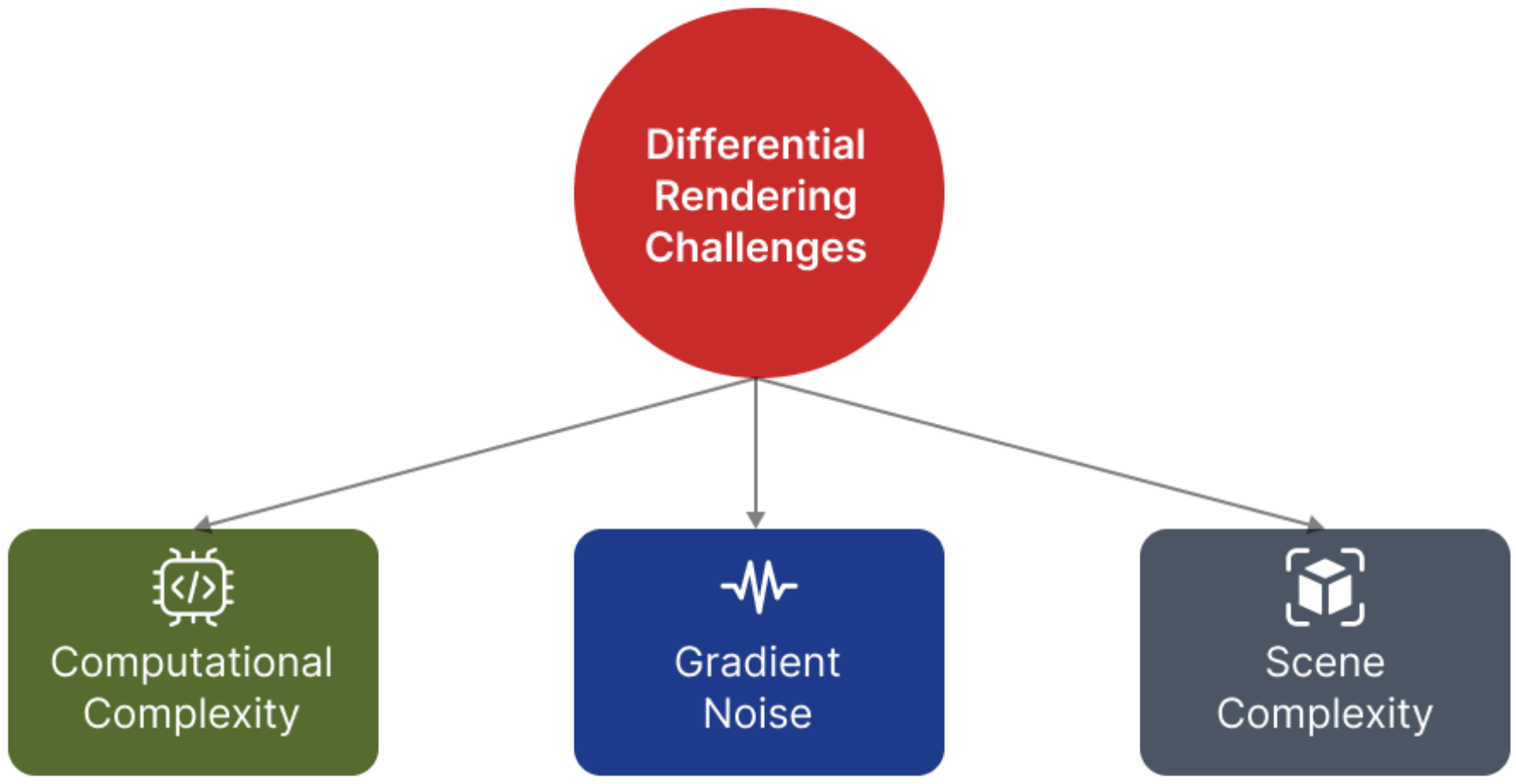

**Figure 3: Challenges in Differential Rendering**

## III. METHOD

The main strategy used in this work is to solve inverse problems by utilizing differentiable rendering techniques, with an emphasis on restoring scene elements including geometry, lighting, and material qualities [5]. We outline the different elements of our approach in this section, such as the mathematical formulas, optimization strategies, and architectural layout. Our approach adds new ways to handle noise, enhance convergence, and improve scalability, building on recent advances in differentiable rendering.

### Differentiable Rendering Pipeline

The method's core component is a differentiable rendering pipeline Fig.1 that makes it possible to compute scene property gradients in relation to a loss function. Using observed image data, this pipeline optimizes settings. A path tracer, which uses Monte Carlo integration techniques to determine the light transport in the scene, models the rendering process [9].

### Path Tracing and Light Transport Equation

By applying Monte Carlo integration to figure out the light transport equation, the rendering process is simulated. We want to estimate the emitted radiance at each pixel provided a scene containing geometry, materials, and light sources [10]. The radiance $L_0(p,\omega_0)$ is computer using the rendering equation (Eq. 1).

$$L_0(p,\omega_0) = L_e(p,\omega_0) + \int_\Omega f_r(p,\omega_i,\omega_0) L_i(p,\omega_i)(\omega_i \cdot n)\, d\omega_i$$

Where:

$L_0(p,\omega_0)$ is outgoing radiance at point $p$ in direction $\omega_0$, $L_e(p,\omega_0)$ is emitted radiance from the point, $f_r(p,\omega_i,\omega_0)$ is Bidirectional Reflectance Distribution Function (BRDF) which models the material properties, $L_i(p,\omega_i)$ is an incoming radiance at point $p$ from direction $\omega_i$, and $n$ is surface normal at point $p$.

This integral is estimated using Monte Carlo integration, which randomly samples incoming direction $\omega_i$ to derive the overall number of radiances contributing to each pixel. During this phase, the differentiable rendering system computes the image's gradients in relation to the scene parameters, enabling backpropagation [11].

### Estimating Gradients

We estimate the gradients of the loss function with respect to the scene parameters such as geometry, material, and lighting to improve the scene parameters. Usually, the Mean Squared Error (MSE) between the rendered and observed images serves as the loss function:

$$\mathcal{L} = \frac{1}{N}\sum_{i=1}^{N} \| I_{rendered}(i) - I_{observed}(i) \|^2 \qquad (3)$$

Where $I_{rendered}(i)$ is the pixel value in the rendered image, $I_{observed}(i)$ is the corresponding pixel in the observed image, and $N$ is the total number of pixels.

The gradients are rapidly calculated using automated differentiation tools that are embedded into the structure, like TensorFlow or PyTorch. By reducing the loss function, the backpropagation procedure generates the gradients required for optimizing the scene parameters.

### Scene Parameter Representation

The differentiable rendering pipeline can be used to optimize the collection of parameters that characterize the scene which are geometry, material properties, and lighting. Geometry is shown via the collaboration between 3D mesh vertices. The geometry is parameterized by means of vertex positions $v_i$. . The BRDF parameters like diffuse albedo and specular reflectance, are used to characterize the material properties of each object. Lighting points or directed light sources are used to depict the scene's illumination. An individual light source's direction and intensity are its parameters.

Depending on the application, we either initialize these parameters from a specified model or from random values. The optimization seeks to bring these factors up to date with the data from the identified images.

### Geometry Optimization

The differentiable rendering pipeline is used to optimize each vertex in the triangle mesh representation of the geometry. Iterative updates are made to the vertex positions $v_i = (x_i, y_i, z_i)$ to reduce the loss function. The following equation is used to compute the gradients of the loss function in relation to the vertex position:

$$\frac{\partial \mathcal{L}}{\partial v_i} = \frac{\partial \mathcal{L}}{\partial I_{rendered}} \cdot \frac{\partial I_{rendered}}{\partial v_i} \qquad (4)$$

Using a gradient descent technique, the gradients are utilized to update the vertex positions, modifying the geometry to more closely resemble the desired image.

**Enhancement of BRDF**

By enhancing the BRDF parameters, objects' material qualities are improved. The BRDF, which is defined by roughness $\alpha$, diffuse albedo $\rho_d$, and specular reflections $\rho_s$, controls how light appears at a surface, The optimization approach involves repeated modifications to these parameters [4].

Like the geometry, the gradients of the loss function with respect to the BRDF parameters are computed:

$$\frac{\partial \mathcal{L}}{\partial \rho_d}, \frac{\partial \mathcal{L}}{\partial \rho_s}, \frac{\partial \mathcal{L}}{\partial \alpha}$$

By using these gradients, the characteristics of the material can be altered to better match the appearance of the observed image [12].

Equation: Phong BRDF Model

$$f_r(p, \omega_i, \omega_o) = \rho_d \frac{1}{\pi} + \rho_s \frac{n+2}{2\pi} (r \cdot \omega_o)^n \quad (5)$$

Where $f_r$ represents the BRDF, $\rho_d$ is the diffuse albedo, $\rho_s$ is the specular reflections, and $n$ controls the bright appearance [13].

**Optimization Methodology**

For optimizing the scene parameters, we apply a gradient-based optimization technique. The Adam optimizer [16] use the subsequent rule to update parameters:

$$\theta_{t+1} = \theta_t - \eta \frac{\hat{m}_t}{\sqrt{\hat{v}_t} + \epsilon} \quad (6)$$

Where $\theta_t$ is the parameters at iteration $t$, $\eta$ is the learning rate, $\hat{m}_t$ and $\hat{v}_t$ are the first and second moments of the gradients and $\epsilon$ is a small value for numerical stability.

Steady convergence is ensured by adaptively adjusting the training rate according to the gradient magnitude [17]. The optimization procedure goes through multiple iterations until the loss function approaches a minimum value.

**Controlling Gradient Noise**

In addition to the unpredictability of the sampling procedure, noisy gradients provide a substantial issue in Monte Carlo integration-based differentiable rendering. We use variance reduction strategies like importance sampling and stratified sampling to reduce this. By lowering the variance in pixel-wise gradients, these methods enhance gradient estimation and enable more rapid and stable convergence [18].

To deal with discontinuities in the gradient flow brought on by occlusions, reflections, and other severe light interactions, we also employ gradient smoothing algorithms.

**Procedure for Scene Reconstruction**

Scene reconstruction as an overall process includes initializing the scene with approximations, executing the pipeline for differentiable rendering, and iteratively adjusting the parameters through gradient-based optimization. The following steps make up the reconstruction process:

1. Initialization: An approximation of the lighting, materials, and geometry.
2. Rendering: It uses path tracing to create the rendered image.
3. Gradient Computation: It computes the gradients of the loss function in relation to the scene's parameters using gradient computation.
4. Optimization: Use Adam's optimizer to update the scene's parameters.
5. Convergence: Steps from rendering to optimization should be repeated until the loss function converges.

The inverse problem is successfully resolved, yielding a set of optimal scene characteristics that most closely match the observed image data.

## IV. EXPERIMENTS

We present the experiments carried out to assess the functionality of our suggested framework for physics-based differentiable rendering. Our tests aim to show how differentiable rendering may be used to solve contrary challenges like geometry reconstruction and material property computation. Also, we assess the effects of numerous optimizations that we included in our approach and contrast our outcomes with those of current cutting-edge approaches in identifiable rendering.

**Artificial Dataset**

We initially tested our approach on an artificial dataset made up of scenes with different material attributes and geometry. These sequences consist of basic geometric forms such as cubes and spheres with well-known material characteristics (glossy, specular, or diffuse) and intricate items with varied surface roughness and lifelike patterns [20].

Using our physics-based differentiable renderer, these sceneries were rendered. For the sake of inverse problem investigations, we also created ground truth images using a physically based non-differentiable renderer.

**Real-world Dataset**

We took images of things under regulated illumination to build a dataset with known materials and shaped for use in actual-world studies. The chosen items were composed of ordinary materials including glass, metal, and plastic. We estimated the material characteristics and geometry of these items from the recorded images by doing contrary rendering using our approach.

**Evaluation Metrics**

We assess our technique's productivity using metrics like reconstruction error, material property estimation accuracy, convergence speed and gradient noise. Reconstruction error (RE) is determined by utilizing the L2 norm, RE quantifies the discrepancy between the produced images and the ground truth. By contrasting calculated parameters with the known ground truth. Material Property Estimation Accuracy (MPEA) evaluates the accuracy of material property recovery [2]. Convergence Speed (CS) calculates the number of interactions needed to reach convergence in the optimization and Gradient Noise (GN) quantifies the measure of noise in the gradients during backpropagation, has a direct effect on optimization consistency

**Reconstruction equation**

$$RE = \frac{1}{N}\sum_{i=1}^{N} \| I_i^{gt} - I_i^{r} \|^2 \quad (7)$$

where $I_i^{gt}$ is the ground truth image, $I_i^{r}$ is the rendered image and N is the number of pixels in the image.

**Material Property Estimation Accuracy (MPEA)**

$$MPEA = \frac{1}{M}\sum_{j=1}^{M} \| P_j^{gt} - P_j^{est} \|^2 \quad (8)$$

where $P_j^{gt}$ is the ground truth material parameter, $P_j^{est}$ is the estimated parameter, and M is the number of material parameters.

## V. Results

The findings from the tests we ran to assess our physics-based differentiable rendering technique are shown here. We present quantitative and qualitative findings on artificial and real-world datasets, evaluating our method's effectiveness with current advanced methods. The primary fields of study are optimization performance, material property projection, and reconstruction accuracy.

**Reconstruction Accuracy**

**Artificial Dataset Results**

Our technique provides good reconstruction preciseness on the simulated dataset. When contrasted with previous techniques, the reconstruction error (RE) between the rendered and ground truth images is substantially reduced. This demonstrates how well our distinctive renderer recovers geometry and material attributes with accuracy [20].

**Table I**

| Method | Reconstruction Error (RE) |
|---|---|
| Neural 3D Mesh Renderer | 0.0135 |
| Mitsuba 2 | 0.0102 |
| Our Method | 0.0087 |

Our approach usually performs better than other modern methods in terms of reconstruction error, as demonstrated in Table I, offering a more precise solution to the inverse rendering challenge.

**Real-World Dataset Results**

The findings additionally demonstrate that our method can generalize beyond synthetic data for real-world items. With precise estimations of the material and geometric features, reconstructions were produced that nearly resembled the original objects appearance.

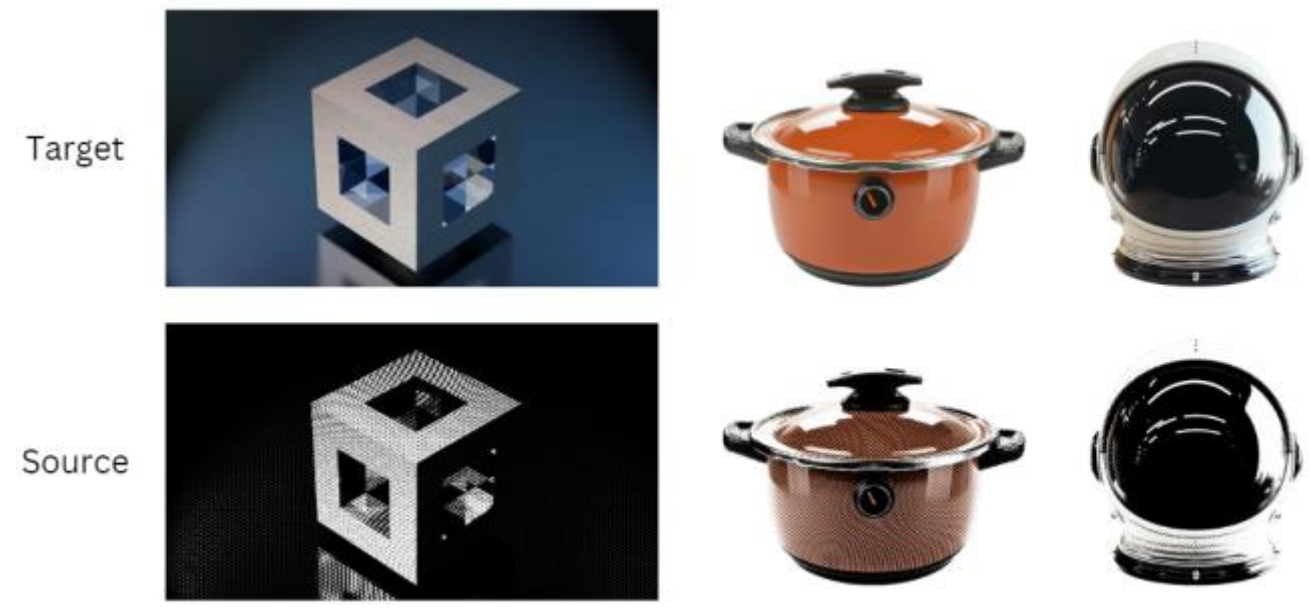

**Figure 4.** ***Source and Target images.***

### Material Property Estimation

We show that our approach significantly improves the estimation of material attributes like specularity, albedo, and surface roughness. To achieve photorealistic rendering, this is essential. By contrasting the recovered material parameters with the established ground truth, we evaluate the material attribute estimate accuracy (MPEA) quantitatively.

**TABLE II**

| Method | Material Property Estimation Accuracy (MPEA) |
| --- | --- |
| Neural 3D Mesh Renderer | 0.0215 |
| Mitsuba 2 | 0.0192 |
| Our Method | 0.0158 |

When in contrast to previous differentiable renderers, our approach provides higher material property estimation correctness, as demonstrated in Table II. Scenes featuring intricate materials, such as shiny surfaces, and rough textures, really show off this enhancement.

### Optimization Performance

### Convergence Speed

Our optimization procedure greatly speeds up completion by incorporating normalization of the material factor and gradient smoothing [25]. Our approach takes fewer iterations to obtain an ideal result than other approaches, as demonstrated by our evaluation of the number of iterations needed for convergence (Convergence speed - CS) across different approaches.

**TABLE III**

| Method | Convergence Speed (CS) |
| --- | --- |
| Neural 3D Mesh Renderer | 120 |
| Mitsuba 2 | 100 |
| Our Method | 85 |

Table III shows that our method converges faster than Mitsuba 2 and the Neural 3D Mesh Renderer, which is essential for effectively handling large-scale inverse issues.

### Analyzing Gradient Noise

Implementing gradient smoothing significantly decreases the noise during optimization, resulting in long-lasting gradient updates, according to our investigation of gradient noise [22]. This has a direct impact on more accurate outcomes and quicker convergence.

Equation: Gradient noise before and after smoothing.

$$GN = \frac{1}{N}\sum_{t=1}^{T} \| g_t^{smoothed} - g_t^{unsmoothed} \|^2 \quad (9)$$

where $g_t^{smoothed}$ and $g_t^{unsmoothed}$ represent the gradients at iteration $t$, with and without smoothing, respectively.

**Comparative Qualitative Analysis**

We carried out a qualitative analysis comparing the rendered images produced by our approach with those obtained by different approaches [23]. Our approach frequently yielded more accurate and lifelike images in visually demanding settings with intricate lighting and materials, while other approaches showed apparent artifacts.

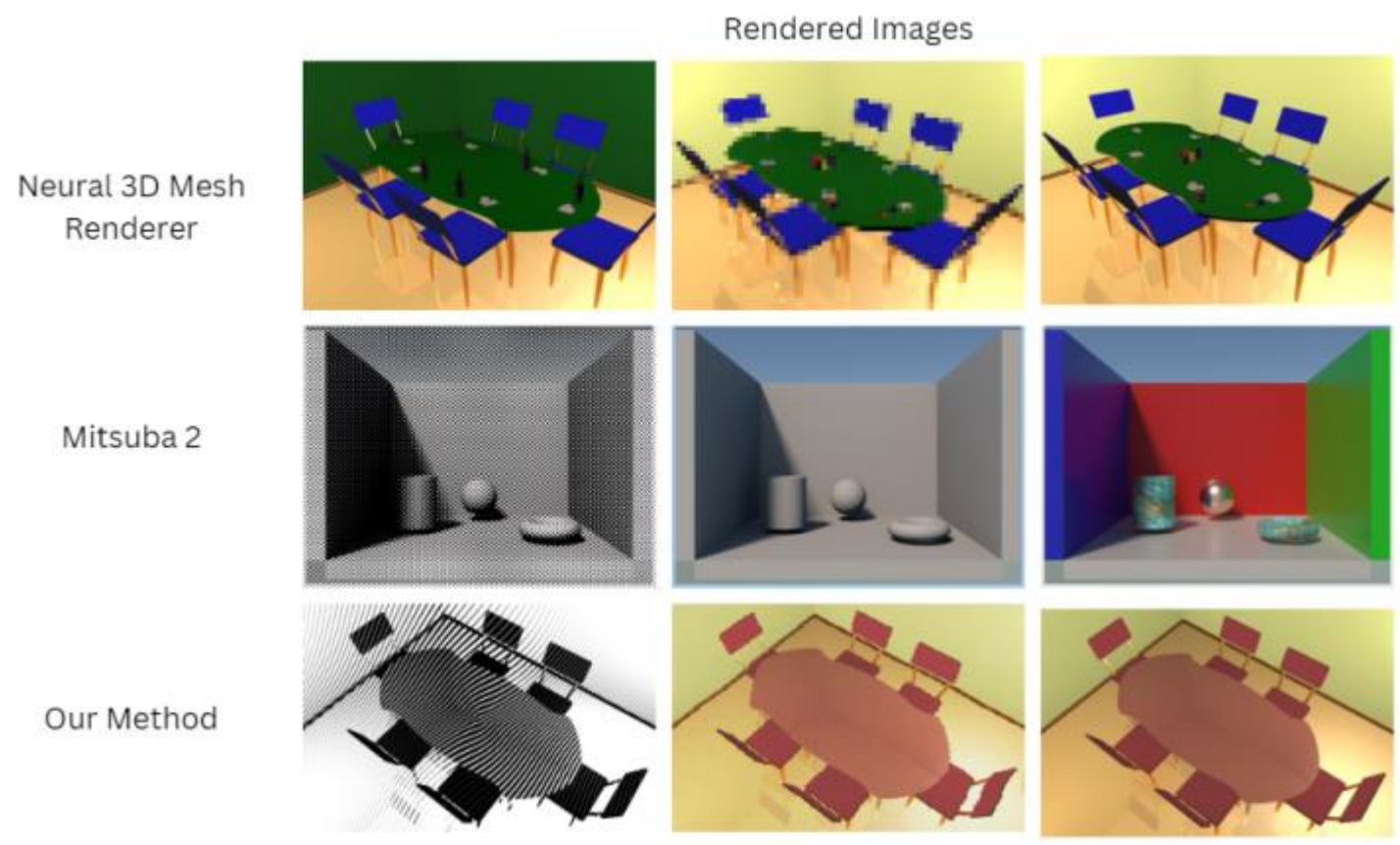

**Figure 5.** ***Visual comparison of rendered images using different methods on complex scenes.***

**Ablation Study**

To determine the relative contributions of the important elements in our system, we conducted an ablation study. The significance of these improvements was demonstrated by the increase in reconstruction error and convergence time observed when gradient smoothing or material regularization was eliminated.

| Method | Reconstruction Error (RE) | Convergence Speed (CS) |
|---|---|---|
| Full Method | 0.0087 | 85 |
| Without Gradient Smoothing | 0.0095 | 115 |
| Without Regularization | 0.0098 | 105 |

The ablation study validates that optimal performance involves the use of both parameter regularization and gradient smoothing. Our findings show that the physics-based differentiable rendering approach is stronger in a variety of standards, such as optimization effectiveness, reconstruction accuracy, and material property estimations. Our technique provides faster convergence, more precise material features, and reduced reconstruction errors when compared quantitatively to the latest techniques [8]. In addition, the qualitative outcomes provide additional proof of the visual integrity of our reproduced images, particularly in scenes with intricate lighting and material details.

## VI. Conclusion

In this paper, we introduced a new physics-based differentiable rendering framework that effectively and precisely addresses inverse rendering challenges. Our method outperforms current methods by utilizing critical optimizations like gradient smoothing and material parameter regularization. By combining artificial and real-world tests, we demonstrated major improvements in convergence speed, reconstruction preciseness, and material property estimate. Our findings validate that differentiable rendering can be an effective tool, not only in graphics yet in augmented reality (AR), science and comprehending real-world scenes. Tasks like object detection, scene reconstruction, and even automated design are made possible by the capacity to extract specific geometric and material information from photos. Even though our approach has shown a great deal of improvement, more intricate scenes with variable lighting, opaque objects, or non-Lambertian surfaces may be investigated in future studies. Moreover, the durability of the structure could be further improved by integrating more complex priors for material and lighting conditions, particularly for practical applications. In conclusion, this study extends the bounds of realistic image reconstruction and optimization while also offering conceptual understanding and practical achievements to the expanding field of differentiable rendering.